\documentclass{article}
\usepackage{spconf,amsmath,graphicx}
\usepackage[detect-all]{siunitx}
\usepackage{mathtools}
\usepackage{comment}
\usepackage{multirow}
\usepackage[linesnumbered,ruled,vlined]{algorithm2e}
\usepackage{bbm}
\usepackage[subrefformat=parens]{subfig}
\usepackage{threeparttable}
\usepackage{booktabs}
\usepackage{amssymb}
\usepackage[hidelinks]{hyperref}

\SetCommentSty{mycommfont}

\makeatletter
\patchcmd{\@algocf@start}
  {-1.5em}
  {0pt}
{}{}
\makeatother

\usepackage{xspace}
\makeatletter
\DeclareRobustCommand\onedot{\futurelet\@let@token\@onedot}
\def\@onedot{\ifx\@let@token.\else.\null\fi\xspace}

\makeatother

\def\equationautorefname~#1\null{(#1\null)}

\newcommand{\vect}[1]{\mbox{\boldmath $#1$}}

\DeclareMathOperator*{\argmax}{arg\,max}
\def\appendixautorefname~#1\null{~#1 \null}

\makeatletter
\newcommand{\figcaption}[1]{\def\@captype{figure}\caption{#1}}
\newcommand{\tblcaption}[1]{\def\@captype{table}\caption{#1}}
\makeatother

\title{Streaming Active Learning for Regression Problems Using~Regression~via~Classification}
\name{%
Shota Horiguchi \qquad
Kota Dohi \qquad
Yohei Kawaguchi%
}
\address{Hitachi, Ltd., Japan}

\begin{document}
\ninept

\maketitle
\begin{abstract} 
One of the challenges in deploying a machine learning model is that the model's performance degrades as the operating environment changes.
To maintain the performance, streaming active learning is used, in which the model is retrained by adding a newly annotated sample to the training dataset if the prediction of the sample is not certain enough.
Although many streaming active learning methods have been proposed for classification problems, few efforts have been made for regression problems, which are often handled in the industrial field.
In this paper, we propose to use the regression-via-classification framework for streaming active learning for regression.
Regression-via-classification transforms regression problems into classification problems so that streaming active learning methods proposed for classification problems can be applied directly to regression problems.
Experimental validation on four real data sets shows that the proposed method can perform regression with higher accuracy at the same annotation cost.
\end{abstract}
\section{Introduction}
Although machine learning models can achieve high prediction performance in a well-controlled environment, they may not be as accurate as expected when actually applied, or their accuracy may deteriorate over time.
Reasons for this include the fact that the initial training data does not adequately cover the operational environment, or that the operational environment is unstable and continues to change from moment to moment, i.e., concept drift \cite{widmer1996learning}.
In such a situation, it is necessary to update the model from time to time using data that come in one after another to maintain the prediction accuracy of the model \cite{gama2014survey,lu2018learning}.

One of the problems in updating a model while operating it is how to obtain samples to be used for updating \cite{baena2006early,bifet2007learning,song2019fuzzy}.
This is not so much of a problem when the groundtruth labels can be obtained immediately, as in the case of time-series prediction of stock prices \cite{harries1995detecting,cavalcante2015approach} or electricity consumption \cite{harries1999splice}.
However, when human measurement and annotation are required to obtain the groundtruth labels, the cost is often an obstacle.

Active learning is a technique for selecting useful samples for training to improve model performance with the lowest possible labeling cost.
There are two types of active learning: pool-based approaches \cite{wu2018pool,wu2019active,traganitis2020active,lin2022pool,bemporad2023active} and streaming approaches \cite{vzliobaite2013active,narr2016stream,krawczyk2017online,krawczyk2018combining}.
In a pool-based approach, the process of extracting a batch of valuable samples from a pre-prepared pool of samples and updating the model is repeated.
This is often accomplished by preparing a utility evaluation function and comparing its value among samples.
In a streaming approach, each time a sample arrives, a decision is made on whether or not to label and update the model.
Streaming active learning is a more difficult problem than pool-based active learning because it requires an evaluation function that can output meaningful absolute values rather than relative values.

Several existing studies have addressed streaming active learning, especially for classification problems \cite{chu2011unbiased,vzliobaite2013active,krawczyk2017online,krawczyk2018combining,castellani2022stream,liu2023online}.
In the classification problem, the maximum value of the probability of belonging to each class output by the classifier can be used as the utility.
The value is bounded by $\left[\frac{1}{K},1\right]$ where $K$ is the number of classes and is easy to use as an absolute utility.
However, similar studies in regression problems have rarely been undertaken.
This is because there is no easy-to-use evaluation function like class-wise probability in regression problems, and absolute evaluation is difficult since the output of the model can take arbitrary values.
The few existing methods even have drawbacks, such as being only applicable to specific regression models, e.g., fuzzy systems \cite{lughofer2017online,ferdaus2019palm} and linear models \cite{sabato2014active,cacciarelli2022stream,chen2022online}.

In this paper, we propose a streaming active learning method for regression problems using regression via classification (RvC) \cite{torgo1997regression}, which transforms a regression problem into a multi-class classification problem.
By using RvC, the target variables are no longer real-valued but discrete classes, and thus arbitrary classifiers can be used to solve regression problems.
This approach is simple but effective, and at the same time, it is useful in that it allows various methods that have been proposed for streaming active learning for classification problems to be applied directly to regression problems.
We demonstrate the effectiveness of the proposed method by evaluating it on a variety of real-world datasets.

\section{Methodology}
\subsection{RvC for Utility Estimation}
Before moving on to the use of RvC \cite{torgo1997regression} for utility estimation in streaming active learning, we first introduce the RvC framework. 
Regression is a task to find a function $f_{\text{reg}}:\mathbb{R}^D\rightarrow\mathbb{R}$ that can predict a real-valued target variable $y\in\mathbb{R}$ given input variables $\vect{x}\in\mathbb{R}^D$:
\begin{equation}
    \hat{y}=f_\text{reg}\left(\vect{x}\right).
\end{equation}
RvC is a framework to solve a regression problem using a classifier.
In the RvC framework, the target variable $y$ is first discretized into $K$ classes using a discretization function $h:\mathbb{R}\rightarrow\left\{1,2,\dots,K\right\}$ and then a classifier that predict a belonging class $g_\text{cls}:\mathbb{R}^D\rightarrow\left[0,1\right]^{K}$ is trained.
For discretization, we used $K$-means clustering in this study.
During inference, the belonging class of input $\vect{x}$ is first estimated and the estimation is then transformed into a regression value as follows:
\begin{align}
    \vect{p}&=g_\text{cls}\left(\vect{x}\right)\label{eq:classification},\\
    \hat{y}&=h^{-1}\left(\vect{p}\right)\label{eq:inverse_discretization},
\end{align}
where $\vect{p}\coloneqq\left[p_1,p_2,\dots,p_K\right]$ is a posterior probability of each class given input $\vect{x}$.
$h^{-1}\left(\cdot\right)$ is an inverse transform of the discretization function $h$ and such a function can be implemented, for example, as a function that returns a mean or median value of the most possible class $\hat{k}=\argmax_k p_k$ in the training set.

Since RvC uses a classifier in \autoref{eq:classification}, the confidence value can be used as a certainty in the same manner in the conventional studies on classification \cite{lewis1994heterogeneous}.
We propose to use the intermediate product of RvC for the utility (or uncertainty) of the sample $\vect{x}$ in regression:
\begin{equation}
    u'\left(\vect{x};K\right)=1-\max_{k\in\left\{1,2,\dots,K\right\}} p_k.
\end{equation}
One problem here is that the optimal number of classes for RvC is not obvious.
A small number of classes results in accurate classification in \autoref{eq:classification} but causes a large discretization error; in contrast, a large number of classes results in low accuracy in classification but the potential discretization error becomes small.
To ease this dilemma, we simply vary the number of classes for $K_1, K_2, \dots, K_N$ and average uncertainties calculated using each classifier. The proposed utility estimator is defined as follows:
\begin{equation}
    u\left(\vect{x}\right)=\frac{1}{N}\sum_{n=1}^{N}u'\left(\vect{x};K_n\right).
\end{equation}

With this technique, various methods of active learning developed for classification problems based on uncertainty \cite{vzliobaite2013active,kottke2015probabilistic,kottke2016multi} can be directly applied to regression problems.
The streaming active learning methodology using the proposed utility is described in the following subsection.

\subsection{RvC-based Utility for Streaming Active Learning}

\begin{algorithm}[t]
    \SetAlgoLined
    \DontPrintSemicolon
    \caption{Streaming active learning framework.}\label{alg:online_active_learning}
    \SetAlgoVlined
    \SetKw{In}{in}
    \SetKw{Continue}{continue}
    \SetKwComment{Comment}{$\triangleright$\ }{}
    \KwIn{initial labeled data $\mathcal{D}=\left\{\left(\vect{x}_m, y_m\right)\right\}_{m=1}^M$, regression model $f_\text{reg}\left(\vect{x}\right)$, utility estimator $g_\text{util}\left(\vect{x}\right)$, budget $b\in\left(0,1\right)$, window size to control budget $w>0$, budget manager $B\left(u, b, w\right)$}
    \KwOut{the estimated value $\hat{y}$ for each input $\vect{x}$}
    \BlankLine
    $f_\text{reg}\mathtt{.train\left(\mathcal{D}\right)}$\label{algline:model_training}\\
    $g_\text{util}\mathtt{.train\left(\mathcal{D}\right)}$\label{algline:utility_training}\\
    \While{sample $\vect{x}$ arrives}{
        $\hat{y}\leftarrow f_\text{reg}\left(\vect{x}\right)$\\
        $u\leftarrow g_\text{util}\left(\vect{x}\right)$\Comment*[r]{Estimate utility of input}\label{algline:utility}
        \If{$B\left(u,b,w\right)~\text{is}~\mathtt{TRUE}$}{\label{algline:decision}
            Acquire the groundtruth label $y$ of $\vect{x}$\label{algline:update_start}\\
            Update $\mathcal{D}$\Comment*[r]{Add $(\vect{x}, y)$ and drop old samples}
            $f_\text{reg}\mathtt{.train}\left(\mathcal{D}\right)$\\
            $g_\text{util}\mathtt{.train}\left(\mathcal{D}\right)$\label{algline:update_end}\\
        }
    }
\end{algorithm}

\autoref{alg:online_active_learning} shows the streaming active learning framework for regression used in this study, which is the simplified version of that used in the previous study \cite{vzliobaite2013active}.
Also, in this study, we trained a regression model separately from a utility estimator based on RvC, i.e., regression using RvC in \autoref{eq:inverse_discretization} is not used (lines \ref{algline:model_training} \& \ref{algline:utility_training}).
This allows us to consider only the effect of sample selection and not the effect of the regression method itself.

Given an input sample $\vect{x}$, its utility, i.e., how much the sample is informative to improve the model's performance if it is labeled and added to the training dataset, is estimated (\autoref{algline:utility}).
The budget manager $B\left(u,b,w\right)$ decides whether to query the groundtruth label of $\vect{x}$ based on the estimated utility within a predetermined budget $b$ (\autoref{algline:decision}).
For example, if $b=0.2$, it is allowed to label about \SI{20}{\percent} of incoming samples.
If the budget manager decides to label the sample, the regression model and utility estimator are retrained using the newly labeled samples (lines~\ref{algline:update_start}--\ref{algline:update_end}).

\section{Experiments}
\subsection{Detailed Implementation of Baselines and Proposed Methods}
For the baselines, we used the random and query-by-committee (QBC) strategies, each of which is detailed below.
\begin{description}
    \item[Random:] In the random strategy, the utility $u$ is sampled from a uniform distribution $\mathcal{U}\left(0,1\right)$ regardless of the input $\vect{x}$. The groundtruth label for the sample is queried if $1-u$ is less than a predetermined budget $b$.
    \item[QBC:] QBC is a powerful baseline for pool-based active learning \cite{zhan2021comparative} and also showed a good performance in streaming settings for classification tasks \cite{krawczyk2017online}.
    QBC uses a diversity of results from $L\left(>1\right)$ models as an indicator of utility:
    \begin{equation}
        u=\sqrt{\frac{1}{L}\sum_{l=1}^{L}\left(\hat{y}_l-\bar{y}\right)^2},
    \end{equation}
    where $\hat{y}_l$ is the output from the $l$-th model and $\bar{y}=\sum_{l=1}^{L}\hat{y}_l$.
    To control the labeling ratio, we used a balancing incremental quantile filter (BIQF) \cite{kottke2015probabilistic}, which can be used in the case when the utility is not bounded. In this study, we used 10 models ($L=10$), and each model was trained using \SI{90}{\percent} of labeled samples available at each time.
\end{description}

For the proposed RvC-based utility estimation, three streaming active learning algorithms that were originally proposed for classification tasks are used in this study.
We used each method implemented in the \texttt{scikit-activeml} framework \cite{kottke2021scikit}.

\begin{algorithm}[t]
    \SetAlgoLined
    \DontPrintSemicolon
    \caption{Budget management with Variable Uncertainty $B\left(u,b,w,s\right)$}
    \label{alg:varun}
    \SetAlgoVlined
    \SetKwInOut{Initialize}{Initialize}
    \SetKw{In}{in}
    \SetKw{Continue}{continue}
    \SetKwComment{Comment}{$\triangleright$\ }{}
    \KwIn{utility $u\in\left[0,1\right]$, budget $b\in\left(0,1\right)$, window size $w>0$, threshold adjustment parameter $s\in\left(0,1\right]$}
    \KwOut{whether to acquire the label $\lambda\in\left\{\mathtt{TRUE}, \mathtt{FALSE}\right\}$}
    \Initialize{spent labeling cost $\hat{b}\leftarrow 0$, threshold $\theta\leftarrow 1$, and store the last values of each during operation}
    \BlankLine
    \uIf{$\hat{b}<b$} {
    \uIf{$1-u<\theta$} {
        $\theta\leftarrow\theta\left(1-s\right)$\Comment*[r]{Decrease uncertainty region}
        $\lambda\leftarrow\mathtt{TRUE}$\label{algline:decrease_threshold}
    } \Else {
        $\theta\leftarrow\theta\left(1+s\right)$\Comment*[r]{Increase uncertainty region}
        $\lambda\leftarrow\mathtt{FALSE}$\label{algline:increase_threshold}
    }
    } \Else {\label{algline:skip1}
        $\lambda\leftarrow\mathtt{FALSE}$\label{algline:skip2}
    }
    $\hat{b}\leftarrow\frac{(w-1)\hat{b}+\lambda}{w}$\Comment*[r]{Update spent cost}\label{algline:approx_cost}
    \Return $\lambda$
\end{algorithm}
\begin{description}
    \item[Variable Uncertainty (VarUn) \cite{vzliobaite2013active}:] Since the distribution of utilities is unknown, it is difficult to predetermine the optimal threshold that satisfies the desired budget. As shown in \autoref{alg:varun}, the VarUn strategy eases this problem by adjusting the threshold based on whether a sample is labeled or not (lines \ref{algline:decrease_threshold} \& \ref{algline:increase_threshold}). The budget was controlled by simply skipping the labeling step if the spent labeling cost exceeded the desired budget at each time (lines \ref{algline:skip1} \& \ref{algline:skip2}). The spent labeling cost was calculated using the approximate spending estimate method \cite{vzliobaite2013active} (\autoref{algline:approx_cost}). As recommended in the paper, the adjusting parameter was set to $s=0.01$ and the initial threshold was set to $\theta=1$ in this study. The window size was set to $w=256$.
    \item[Split \cite{vzliobaite2013active}:] This is a combination of the random and VarUn strategies. With the predetermined parameter $\nu\in\left(0,1\right)$, at each time, if the random variable $\eta\sim\mathcal{U}\left(0,1\right)$ exceeds $\nu$, the decision whether label $\vect{x}$ is made using the VarUn strategy and otherwise the random strategy is used. The budget was controlled in the same manner as the VarUn strategy. The randomization parameter was set to $\nu=0.5$.
    \item[Probabilistic active learning (PAL) \cite{kottke2015probabilistic,kottke2016multi}:]
    PAL is a method to predict an influence if acquiring a groundtruth label of input sample using its neighborhood. 
    In this paper, we used an extension of multi-class PAL \cite{kottke2016multi} to streaming environments \cite{kottke2015probabilistic}. In PAL, the utility is computed as the gain in accuracy. 
    The budget was controlled using BIQF.
\end{description}

We used support vector machines for the main regressor, regressors in the QBC-based utility estimator, and classifiers in the RvC-based utility estimator.
Note that our method can adopt any type of classifier if it has a way to calculate class-wise probabilities (e.g., the \texttt{predict\_proba} method of most of the classifiers implemented in scikit-learn \cite{pedregosa2011scikit}), rather than conventional methods that limit the classifier type \cite{lughofer2017online,sabato2014active,ferdaus2019palm,cacciarelli2022stream,chen2022online}.

\subsection{Datasets}
For evaluation, we used four real-world datasets below. Each dataset has been used for research on drifting data streams in the literature \cite{song2019fuzzy,oikarinen2021detecting}.

\begin{description}
    \item[House \cite{pace1997sparse}:] This is a dataset of housing value of blocked areas in California from the 1990 census. It contains 20,640 samples and the problem is a regression of median house value using eight features: median income, housing median age, total rooms, total bedrooms, population, households, latitude, and longitude.
    \item[SMEAR \cite{vzliobaite2014regression}:] This is a collection of seven years of 30-min interval meteorological sensor data collected at the SMEAR II station in Finland. The dataset is preprocessed to remove missing values, resulting in 140,576 samples in total.\footnote{\url{https://github.com/zliobaite/paper-missing-values}} The problem is a regression of the ratio of the actual radiation to the theoretical radiation using 36 meteorological features.
    \item[Solar\footnotemark:]\footnotetext{\url{https://www.kaggle.com/datasets/dronio/SolarEnergy}} This dataset stores four months of 5-min interval meteorological sensor data from the HI-SEAS weather station in Hawaii. It contains 32,686 samples and the problem is a regression of solar radiation using five features: temperature, barometric pressure, humidity, wind direction, and wind speed.
    \item[Bike \cite{fanaee2014event}:] This dataset is an hourly bike rental count in Washington, D.C., USA in 2011 and 2012. It contains 17,379 samples and the problem is a regression of rental bike counts using 12 features including seasonal and meteorological ones.
\end{description}

\subsection{Preliminary Results of Offline Evaluation}
We first evaluated the methods in an offline manner to see whether the utility calculated based on the proposed method can be an indicator of error between prediction and groundtruth.
We performed 5-fold cross-validation by splitting each data in sequential order to compare QBC-based and RvC-based utilities.
Since these utility values have different ranges, that is, utilities from QBC can take any value greater than or equal to zero while those from RvC are bounded to $\left[0,1\right]$, we evaluated the relationship between absolute error and rank of utility instead of absolute utility value.

\begin{figure}[t]
    \centering
    \subfloat[][House]{\includegraphics[width=0.49\linewidth]{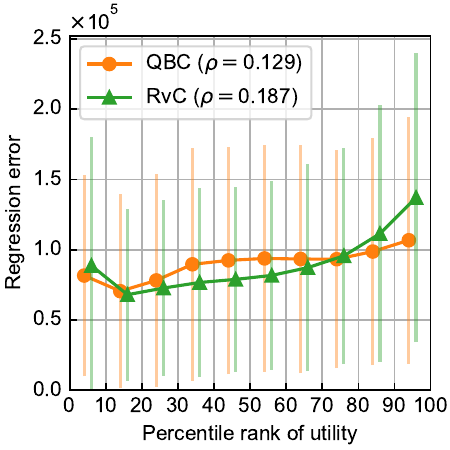}}\hfill
    \subfloat[][SMEAR]{\includegraphics[width=0.49\linewidth]{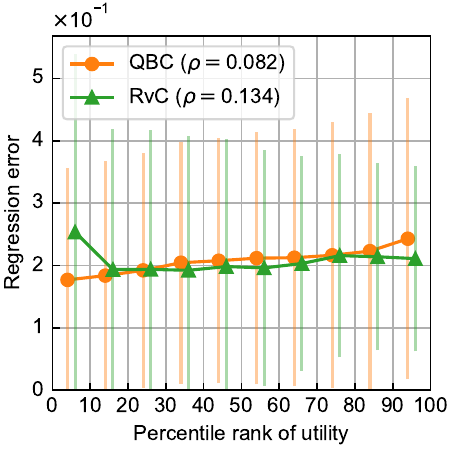}}\\
    \subfloat[][Solar]{\includegraphics[width=0.49\linewidth]{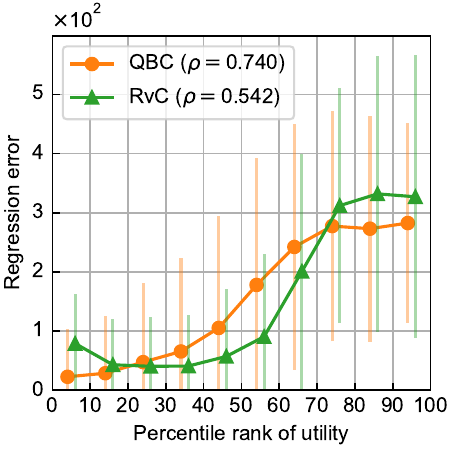}}\hfill
    \subfloat[][Bike]{\includegraphics[width=0.49\linewidth]{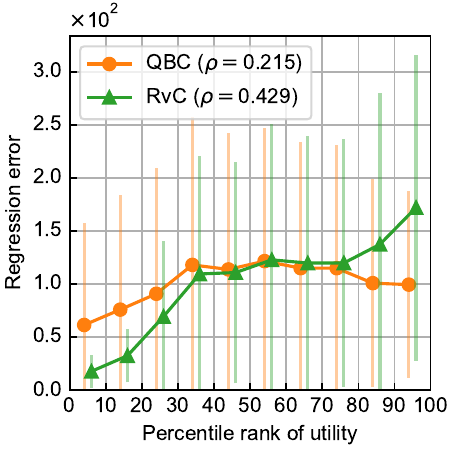}}
    \caption{Binned scatter plots of percentile rank of utility vs. absolute error using QBC-based and RvC-based utility estimators in the offline evaluation. Each error bar represents a standard deviation.}
    \label{fig:offline_results}
\end{figure}

\begin{table*}
    \centering
    \caption{RMSE averaged across different budgets in the streaming evaluation. The numbers in the parentheses are the rank of RMSE among the five methods.}
    \label{tbl:results_streaming}
    \begin{tabular}{@{}ll@{\hskip 0.2em}cl@{\hskip 0.2em}cl@{\hskip 0.2em}cl@{\hskip 0.2em}cl@{\hskip 0.2em}c@{}}
        \toprule
        &\multicolumn{4}{c}{Baseline}&\multicolumn{6}{c}{Proposed}\\\cmidrule(lr){2-5}\cmidrule(l){6-11}
        Dataset&\multicolumn{2}{c}{Random}&\multicolumn{2}{c}{QBC}&\multicolumn{2}{c}{VarUn}&\multicolumn{2}{c}{Split}&\multicolumn{2}{c}{PAL}\\\midrule
        House&$1.0881\times 10^5$&(4)&$1.0943\times 10^5$&(5)&$1.0784\times 10^5$&(3)&$1.0784\times 10^5$&(2)&$1.0780\times 10^5$&(1)\\
        SMEAR&$2.9303\times 10^{-1}$&(4)&$2.9550\times 10^{-1}$&(5)&$2.9020\times 10^{-1}$&(2)&$2.9006\times 10^{-1}$&(1)&$2.9079\times 10^{-1}$&(3)\\
        Solar&$3.6161\times 10^2$&(3)&$3.6439\times 10^2$&(5)&$3.6021\times 10^2$&(2)&$3.6001\times 10^2$&(1)&$3.6176\times 10^2$&(4)\\
        Bike&$1.6734\times 10^2$&(2)&$1.7396\times 10^2$&(5)&$1.6804\times 10^2$&(4)&$1.6784\times 10^2$&(3)&$1.6634\times 10^2$&(1)\\\midrule
        Average rank&\multicolumn{2}{c}{3.25}&\multicolumn{2}{c}{5}&\multicolumn{2}{c}{2.75}&\multicolumn{2}{c}{1.75}&\multicolumn{2}{c}{2.25}\\
        \bottomrule        
    \end{tabular}%
\end{table*}

\autoref{fig:offline_results} shows the binned scatter plot of the percentile rank of utility and absolute error between groundtruth and predicted values.
We also showed Spearman's rank correlation coefficient $\rho$ in each figure.
A gradual increase in regression error with increasing utility can be observed for both QBC and RvC.
Comparing the two methods, the proposed RvC-based utility estimation method outperformed the QBC-based method in all the datasets except for the Solar dataset.
Note that even if QBC and RvC are comparably good utility estimators, the RvC-based one is advantageous in that its utilities allow direct application of the variety of active learning methods proposed for classification problems.

\subsection{Results of Streaming Evaluation}

We next evaluated the baselines and proposed methods in a streaming manner using \autoref{alg:online_active_learning}.
For each dataset, 100 trials are performed and the averaged root mean squared errors (RMSEs) are reported.
In each experiment, i) we randomly extracted 2,100 consecutive samples, ii) trained an initial model using the first fully labeled 100 samples, and iii) evaluated RMSE in a prequential (interleaved test-then-train) manner using the last 2,000 samples.
We also applied a sliding window with a size of 500 and the samples outside the window is omitted in the training; thus, the increase in model training time due to an increase in the number of samples over time is capped by this window width.
The budget was varied by $b\in\left\{0.05, 0.1, 0.15, 0.2, 0.25, 0.3, 0.35, 0.4\right\}$.

\begin{figure}[t]
    \centering
    \subfloat[][House]{\includegraphics[width=0.49\linewidth]{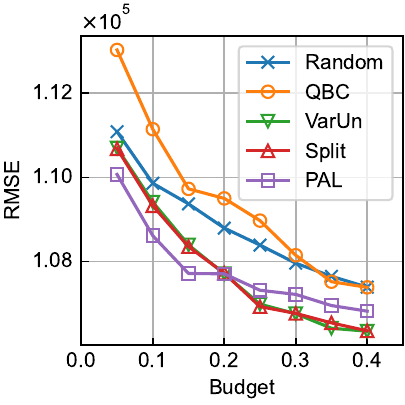}}\hfill
    \subfloat[][SMEAR]{\includegraphics[width=0.49\linewidth]{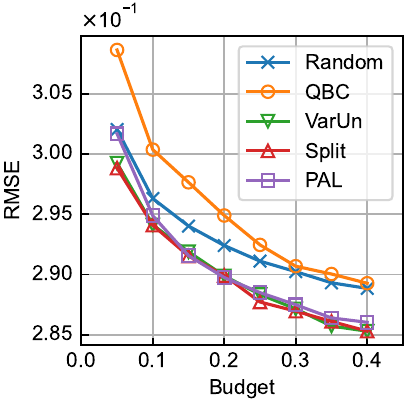}}\hfill\\
    \subfloat[][Solar]{\includegraphics[width=0.49\linewidth]{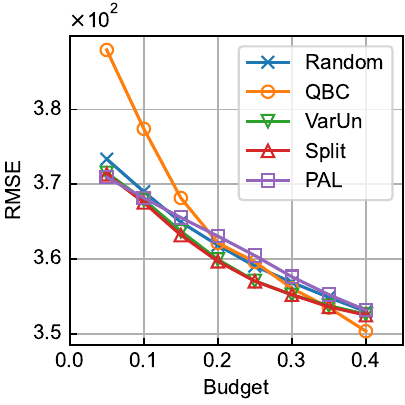}}\hfill
    \subfloat[][Bike]{\includegraphics[width=0.49\linewidth]{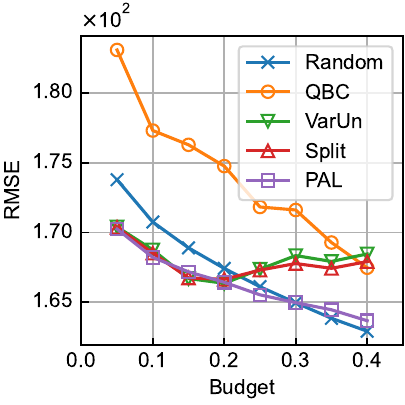}}
    \caption{Budget-wise RMSE in the streaming evaluation.}
    \label{fig:results_streaming}
\end{figure}

\autoref{tbl:results_streaming} shows the averaged RMSE across the varied budget values.
The fact that the average ranks of VarUn, Split, and PAL are smaller than those of random and QBC supported that the proposed RvC-based utility estimation is effective when combined with the conventional streaming active learning strategy for classification tasks.

\autoref{fig:results_streaming} shows budget-wise RMSE on the four datasets.
RvC-based methods outperform the baselines, especially when the budget is small.
Among them, the performance of PAL is consistently good regardless of the budget while VarUn and Split are sometimes degraded, which is consistent with the results in classification tasks \cite{kottke2015probabilistic}.
In conclusion, it can be said that methods developed for classification problems can be directly applied to regression problems by using RvC, and that if a method is superior for classification problems, it will also be superior for regression problems.
RvC has the potential to easily extend the method proposed for classification problems to regression problems in scenarios other than streaming active learning, e.g., drift detection \cite{baena2006early,bifet2007learning}, which is left for future work.

\section{Conclusion}
In this paper, we proposed a utility estimation method for regression problems and demonstrated its effectiveness in streaming active learning.
In the proposed method, we used the RvC framework to convert regression problems into classification problems and thus many conventional methods of streaming active learning for classifications can be directly applied.
Experimental results on four real datasets showed the proposed method improved streaming active learning for regression problems.

\bibliographystyle{IEEEbib}
\bibliography{refs}

\end{document}